\documentclass[runningheads, orivec]{llncs}

\usepackage{graphicx} 
\usepackage{comment}
\usepackage{changepage}
\usepackage{graphicx}
\usepackage{tabularx}
\usepackage{xcolor}
\usepackage{colortbl}
\usepackage{float}
\newcommand{\ignore}[1]{}
\usepackage{siunitx}
\usepackage{changepage}

\usepackage[labelfont=bf]{caption}
\usepackage{subcaption}
\usepackage{afterpage}
\usepackage{pdfpages}

\usepackage{booktabs}
\usepackage{makecell}

\usepackage{amsmath,amssymb}
\usepackage{multirow} 
\usepackage[utf8]{inputenc}
\usepackage{algorithm}
\usepackage{algpseudocode}

\definecolor{cust}{RGB}{84,178,255} 
\usepackage[colorlinks,allcolors=cust, bookmarks=false]{hyperref}
\makeatletter
\newcommand{\printfnsymbol}[1]{%
  \textsuperscript{\@fnsymbol{#1}}%
}
\makeatother

\usepackage{multicol}
\usepackage{needspace}

\usepackage{etoc}   
\usepackage{newfloat}

\usepackage[a4paper, margin=1.67in]{geometry} 

\begin{document}

\title{Context-aware Skin Cancer Epithelial Cell Classification with Scalable Graph Transformers}
\date{december 2025}
\titlerunning{Context-aware classification with Graph Transformers}

\author{Lucas Sancéré \inst{1,2,3,4} \and Noémie Moreau \inst{2,3,4} \and Katarzyna Bozek \inst{2,3,4}} 
\authorrunning{Sancéré et al.}

\institute{ 
    Faculty of Mathematics and Natural Sciences, University of Cologne, Cologne, North Rhine-Westphalia, Germany \\ 
    \and
    Institute for Biomedical Informatics, Faculty of Medicine and University Hospital Cologne, University of Cologne, Cologne, North Rhine-Westphalia, Germany \\ 
    \and
    Center for Molecular Medicine Cologne (CMMC), Faculty of Medicine and University Hospital Cologne, University of Cologne, Cologne, North Rhine-Westphalia, Germany \\ 
    \and
    Excellence Cluster on Cellular Stress Responses in Aging-Associated Diseases (CECAD), University of  Cologne, Cologne, North Rhine-Westphalia, Germany \\ 
    \email{lsancere@uni-koeln.de, k.bozek@uni-koeln.de}
}

\maketitle   

\begin{adjustwidth}{-0.625cm}{-0.625cm} 
\begin{abstract}
Whole-slide images (WSIs) from cancer patients contain rich information that can be used for medical diagnosis or to follow treatment progress. To automate their analysis, numerous deep learning methods based on convolutional neural networks and Vision Transformers have been developed and have achieved strong performance in segmentation and classification tasks. However, due to the large size and complex cellular organization of WSIs, these models rely on patch-based representations, losing vital tissue-level context. We propose using scalable Graph Transformers on a full-WSI cell graph for classification. We evaluate this methodology on a challenging task: the classification of healthy versus tumor epithelial cells in cutaneous squamous cell carcinoma (cSCC), where both cell types exhibit very similar morphologies and are therefore difficult to differentiate for image-based approaches. We first compared image-based and graph-based methods on a single WSI. Graph Transformer models SGFormer and DIFFormer achieved balanced accuracies of $85.2 \pm 1.5$ ($\pm$ standard error) and $85.1 \pm 2.5$ in 3-fold cross-validation, respectively, whereas the best image-based method reached $81.2 \pm 3.0$. By evaluating several node feature configurations, we found that the most informative representation combined morphological and texture features as well as the cell classes of non-epithelial cells, highlighting the importance of the surrounding cellular context. We then extended our work to train on several WSIs from several patients. To address the computational constraints of image-based models, we extracted four $2560 \times 2560$ pixel patches from 93 WSIs of 84 patients and converted them into graphs. In this setting, DIFFormer achieved a balanced accuracy of $83.6 \pm 1.9$ (3-fold cross-validation), while the state-of-the-art image-based model CellViT256 reached $78.1 \pm 0.5$. DIFFormer training was also substantially faster, requiring only 32 minutes for one single cross-validation fold compared with approximately 5 days for CellViT256. Overall, these results suggest that graph-based approaches constitute a promising alternative to traditional computer vision methods for object classification tasks that use larger spatial context for improved performance. Our tool and datasets are available through our github repository: \href{https://github.com/bozeklab/wsi-graph-formers}{https://github.com/bozeklab/wsi-graph-formers}. 
\end{abstract}
\end{adjustwidth}

\section*{Introduction}

Hematoxylin and Eosin (H\&E) staining \cite{Wittekind2003} is widely used in pathology and serves as a standard staining method for tissue examination. The resulting scans, called Whole-Slide Images (WSIs), are high-resolution digital images of entire tissue sections captured from microscope slide. Deep learning has substantially advanced WSI analysis, primarily through convolutional neural networks (CNNs) \cite{Coudray2018,Campanella2019} and more recently Transformer models \cite{Shao2021,Chaurasia2025,Pisula2025}. Due to the large size of WSIs, their automated processing requires first splitting the image into smaller patches, extracting features from the patches, before aggregating them for prediction or segmentation of the entire slide. This approach enables  analysis of the large images, however does not allow the models to capture the tissue structure as a whole as each patch contains only a small area of the image. \medskip

To model tissue spatial organization, recent approaches represent WSIs as graphs, where nodes correspond to image regions and edges encode their spatial relationships. Different graph neural network (GNN) architectures were applied to such graphs for survival prediction outcome\cite{Kipf2017,Chen2021pointcloud,Liu2023,Zhao2023coads}. Notably, Graph Transformer models were used for lung, breast, and kidney cancer WSI classification based on individual patches \cite{Zheng2022,Shi2024,Ramanathan2024}. These models demonstrate the effectiveness of graph representations and GNNs for WSI analysis. While graphs are powerful models for representing WSIs, graphs that are based on patches rather than individual cells do not capture the detailed tissue composition and cellular interactions within it.  \medskip

Graph representation in which nodes correspond to individual cells or nuclei and edges link cells based on their spatial proximity or feature similarity is a rich and detailed model of a tissue sample. HistoCartography toolkit \cite{Jaume2021-HistoCartography} was designed to build such cell graphs but is limited to individual patches of the full WSI. Such graphs were used for cell classification, through node classification with GNNs and notably Graph Transformers, however using only patch-level context \cite{Lou2023structure,Hassan2022,Lou2024cellgraph}. Similarly, hierarchical graphs including cell representations from WSI patches were developed for tissue segmentation, cancer grading, and breast cancer classification  \cite{Javed2020,Zhou2019,Pati2022}. In these approaches nodes represent individual cells enabling biologically meaningful modeling, nonetheless, graphs remain restricted to local patches and do not capture entire WSIs. Achieving tissue-level representation and analysis with graphs requires incorporating all cells within a WSI and developing GNNs and their training strategies capable of operating at this scale. Vanea et al. \cite{Vanea2022} took a first step in this direction by constructing cell graphs spanning entire WSIs of placental tissue, in which each node corresponds to a single cell and edges connect spatially neighboring cells. However, the classification performance of the evaluated GNNs remained limited, particularly for tissue types whose identification depends on context beyond a cell's immediate neighborhood.  \medskip

Classical GNNs rely on localized message passing, which limits their ability to capture long-range dependencies and restricts their scalability to large graphs. Graph Transformer models \cite{Kim2022,Ying2021,Rampasek2022} enable to address this challenge with global attention across nodes while leveraging efficient attention mechanisms. However, the quadratic complexity of the attention mechanism with regard to the input tokens, makes training on large graphs impossible. To overcome this issue new architectures of scalable Graph Transformers with linear complexity were developed such as Nyströmformer \cite{Xiong2021} and more recently NodeFormer \cite{Wu2022nodeformer}, DIFFormer \cite{Wu2023difformer} and  Simplified Graph Transformers (SGFormer) \cite{Wu2023sgformer}. These models were used for node classification on large-graph benchmarks such as \textsf{ogbn-proteins} \cite{Hu2020}, \textsf{Amazon-M2} \cite{Jin2023} or \textsf{pokec} \cite{Takac2012}. Training such models on cell-level WSI graphs could significantly improve node classification accuracy over traditional patch-based methods.   
\medskip

In this work we propose a graph-based approach for tumor cell classification in cutaneous squamous cell carcinoma (cSCC), the second most common non-melanoma skin cancer worldwide \cite{Guo2023,Howell2023}. Previous work showed that distinguishing between healthy and tumor epithelial cells in the cSCC is challenging using classical patch-based segmentation methods. The morphologies of these cells are very similar and only a broader tissue-level context capturing cell spatial organization allows to correctly classify them in these two types \cite{Sancere2026}. We tackle the challenge of  discriminating between healthy and tumor epithelial cells by building large WSI graph and training scalable Graph Transformers with linear complexity. We use a previously published dataset of cSCC WSIs \cite{Sancere2026} and construct graphs at two different levels, WSI and patch, and term these approaches ``WSI-Graph" and ``TILE-Graphs" to compare GNNs and image-based models on epithelial cell classification. We first asses the performance of GNN models on WSI-Graph and find the best combination of node features. We then evaluate image-based and graph-based approach on a several patient configuration with TILE-Graphs. To the best of our knowledge this work is the first to:
\begin{itemize}
    \item[$\bullet$] Apply graphs to improve classification of epithelial cells as healthy or tumor in the cSCC, 
    \item[$\bullet$] Compare graph-based and image-based methods on the same underlying data represented as images and corresponding graph structures.
\end{itemize}

\section*{Methods}

\subsection*{From WSIs to cell graphs}
\label{fromwsi}

To apply a GNN to the classification of healthy versus tumor epithelial cells, we converted a WSI of cSCC into a structured graph representation. In this graph, each node corresponds to a cell nucleus and encodes morphological and texture features, as well as the associated cell class. Edges connect neighboring cells, thereby capturing their spatial relationships. \medskip

To obtain these nuclei and their initial labels, we used cSCC Hovernet \cite{Sancere2026}, a model previously developed for cell segmentation and classification. cSCC Hovernet detects, segments and then categorizes cells into five types: granulocytes, plasma cells, lymphocytes, stromal cells, and epithelial cells (without distinguishing between healthy and tumor cells). Using tumor regions annotated by an expert pathologist, we refined this classification by splitting epithelial cells into two subclasses: tumor epithelial and healthy epithelial. Specifically, epithelial cells located inside annotated tumor regions were relabeled as tumor epithelial, whereas epithelial cells outside these regions were relabeled as healthy epithelial. Importantly, not all cells within a tumor region are tumor cells; only those previously identified as epithelial are reassigned as tumor epithelial. As a result, each cell of the entire WSI is categorized into one of the six classes: granulocytes, plasma cells, lymphocytes, stromal cells, tumor epithelial and healthy epithelial. \medskip

Based on the cell segmentation and classification, we represent a WSI as a graph, where each detected cell nucleus forms a node. Each node $i \in \{1,\dots,N\}$ among all $N$ nodes is associated with a feature vector $\mathbf{h}_i = \left[ \mathbf{f}_i \;\| \; \mathbf{c}_i \right] \in \mathbb{R}^{l + 6}$ where $\mathbf{f}_i \in \mathbb{R}^{l}$ encodes $l$ morphological and texture features, and $\mathbf{c}_i \in \{0,1\}^{6}$ denotes the cell class encoded as a one-hot vector.  We built the undirected, node-attributed graph $G = (\mathbf{H}, \mathbf{A})$, where $\mathbf{H} \in \mathbb{R}^{N \times (l+6)}$ is the feature matrix concatenation of each feature vector $\mathbf{h}_i$, and where $\mathbf{A} = [A_{ij}]\in \mathbb{R}^{N \times N}$ is the adjacency matrix. $[A_{ij}]=1$ if there exists an edge connecting node $i$ and node $j$, otherwise, $[A_{ij}]=0$. We define a threshold distance $r_0$ for node connections and denote $d_{E_{ij}}$ the Euclidean distance between the centroids of the cell nuclei in the WSI corresponding to nodes $i$ and $j$ in $G$. If $d_{E_{ij}}< r_0$, then there is an edge between $i$ and $j$ and $[A_{ij}]=1$. The resulting graph is shown in  \textbf{Figure \ref{fig1}a}. The simplification and splitting steps are described in the following sections.

\subsection*{Graph simplification}
\label{simplification}

We simplify the graph by first removing all nodes corresponding to cell nuclei that were detected by cSCC Hovernet but not classified (because of low confidence). In addition, nodes of epithelial cells with degree zero are more likely to be miss-classified, as epithelial cells are forming a compact group in the tissue. The nodes corresponding to these cells are also removed. \medskip

The graph is subsequently used for binary node classification, where nodes representing epithelial cells are classified as either tumor or healthy. Not all nodes are considered equally during training and inference. In particular, nodes located in the local neighborhood of tumor and healthy epithelial cells are expected to have a greater influence on the classification through message passing than nodes that are further away. To both reduce the computational complexity of the graph during training \ignore{(less RAM and  video RAM needed during training)}and simplify its structure, we remove nodes that are away from epithelial cells by a given number of graph edges. This reduction limits the number of contributing nodes in the aggregation of latent features during graph convolution. 

\medskip

We define nodes corresponding to tumor or healthy epithelial cells as anchor nodes around which the graph will be simplified. We retain all nodes that lie within a geodesic distance $d_G$ (number of edges in a shortest path connecting 2 nodes) of at most $k$ from at least one anchor node. We define $k$-max-hops as the maximum allowed geodesic distance between a cell and its nearest anchor node. We define the class indicator matrix $\mathbf{C} \in \{0,1\}^{N \times 6}$ where the $i$-th row $\mathbf{c}_i \in \{0,1\}^{6}$ encodes the class of node $i$. To specify which class is an anchor, we introduce a binary class-selection vector $ \boldsymbol{s} \in \{0,1\}^{6}$ where $s_{p}=1$ indicates that class $p$ is selected as an anchor class. For each node $i \in \{1,\dots,N\}$ we define the anchor indicator vector $\mathbf{a} \in \{0,1\}^{N}$ as: 
\[
a_i =
\begin{cases}
1, & \text{if } (\mathbf C \boldsymbol{s})_i > 0\\
0, & \text{otherwise}
\end{cases}
\]

The anchor indicator $a_i$ equals 1 if and only if node $i$ belongs to a class selected by $\boldsymbol{s}$ (here tumor and healthy epithelial). We finally introduce the node mask $\mathbf{m^{k}} \in \{0,1\}^{N}$ defined as:
\[
m_i^{k} = 1
\;\Longleftrightarrow\;
\min_{j:\,a_j = 1}
d_G(i,j)
\le k, 
\qquad \forall i \in \{1,\dots,N\}
\]
$d_G$ is computable for any pair of nodes, but interestingly, by using the properties of the powers of the adjacency matrix one doesn't need to directly compute distances. Indeed $A_{ij}^{q}$ is the number of walks of length $q$ from node $i$ to node $j$. From this we can deduce $d_G(i,j)$ as the smallest non negative $q$ such as $A_{ij}^{q}>0$. Another definition of $\mathbf{m}$ is:
\[
m_i^{k} =
\begin{cases}
1, & \text{if}
\left[
\left(
\sum_{q=0}^{k} \mathbf A^{q}
\right)
\mathbf a
\right]_i
> 0\\
0, & \text{otherwise}
\end{cases}
\qquad \forall i \in \{1,\dots,N\}
\]

After simplification of $G$ we obtain the induced subgraph $G^{(k)} = (\mathbf{H^{(k)}}, \mathbf{A^{(k)}})$  with the induced adjacency matrix $\mathbf{A^{(k)}} \in \mathbb{R}^{N \times N}$ and masked features matrix $\mathbf{H^{(k)}} \in \mathbb{R}^{N \times (l+6)}$ given by:
\begin{equation}
\mathbf{M^{(k)}}= \operatorname{diag}(\mathbf{m^{k}}), \ \mathbf{A^{(k)}} = \mathbf{M^{(k)}AM^{(k)}}, \ \mathbf{H^{(k)}} = \mathbf{M^{(k)}H}   
\label{}
\end{equation} 

\subsection*{Implementation details of WSI-Graph}
\label{dataset}

\begin{figure}[]
    {\centering
    \includegraphics[width=0.75\textwidth]{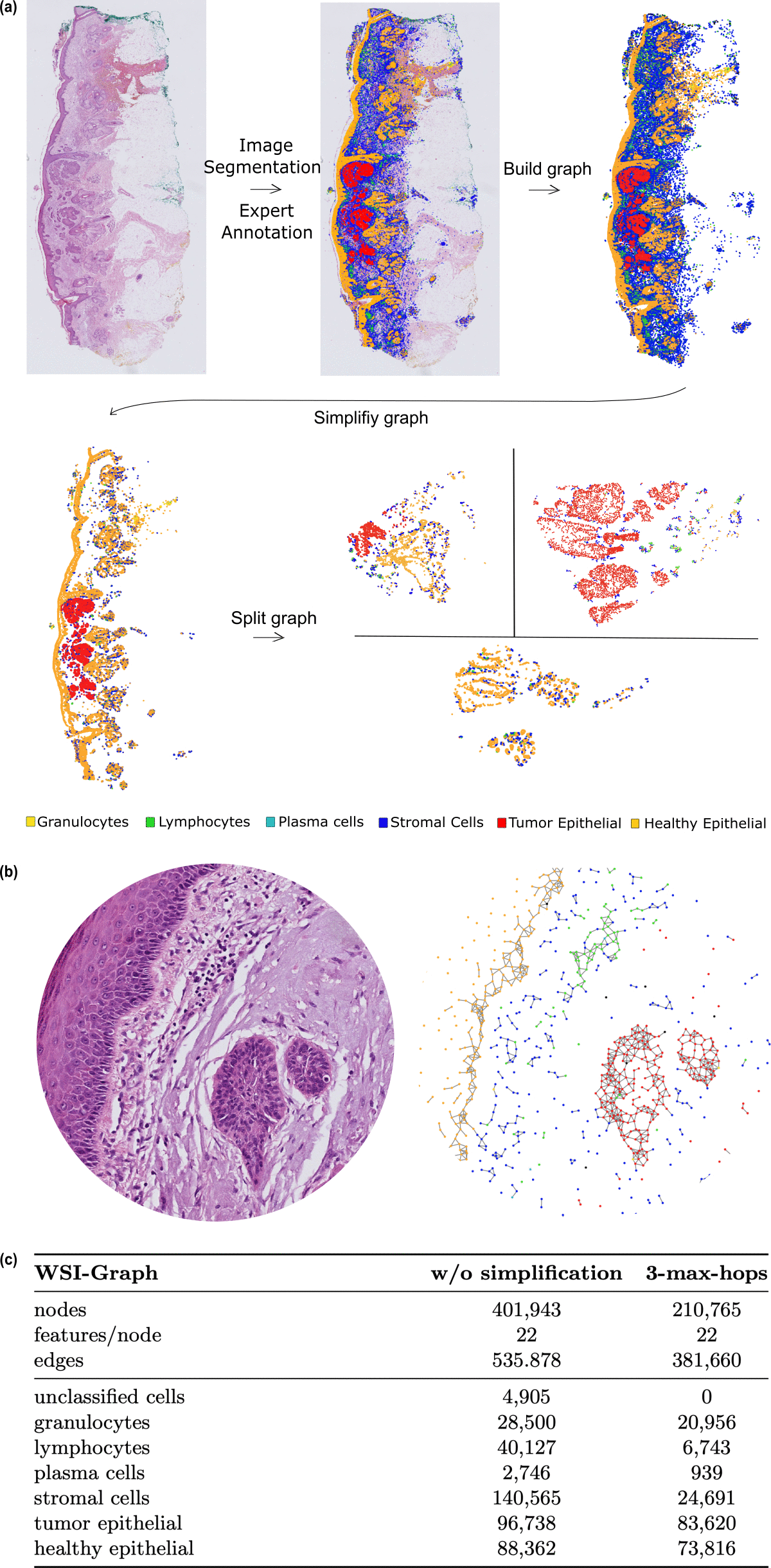}
    \par}
    \caption[Caption Title]{ \small \textbf{Description of WSI-Graph.}
    \footnotesize  \textbf{(a)} Steps to generate  WSI-Graph. First a WSI from cSCC patient is segmented using SCC Hovernet and tumor regions are annotated by an expert to refine segmentation. From this segmentation map we build a graph, simplify it around anchor nodes and optionally split it with K-means on centroid coordinate features. \textbf{(b)} Zoom into the graph, edges generated with threshold distance $r_0=50\operatorname{pixels}$ corresponding to $r_0\approx \SI{11.5}{\micro\meter}$ are shown. \textbf{(c)} Number of edges, nodes, node features and instances of given cell classes before and after simplification (here $k=3$ max-hops simplification).}
    \label{fig1}
\end{figure}

From a WSI of cSCC with annotated tumor regions we build and then simplify a graph of this tissue as described in previous sections. We also split the simplified graph into 100 non-overlapping subgraphs of similar sizes. To build the subgraphs we run K-means algorithm on the 2D coordinates of nuclei to form K spatial clusters. We use these subgraphs to evaluate GNN's performance on node binary classification task. The process of graph building is depicted in \textbf{Figure \ref{fig1}a} where each node is colored according to its cell class, and spatially located using nucleus centroid coordinates.  \textbf{Figure \ref{fig1}b} shows the graph edges. The threshold distance for edges is $r_0=50\operatorname{pixels}$ corresponding to $r_0\approx \SI{11.5}{\micro\meter}$. This threshold ensures that neighboring nuclei within cells that are likely to interact or form structures together are connected by an edge, and nuclei that are far apart are not.  \medskip 

A comprehensive description of WSI-Graph before and after simplification is shown in \textbf{Figure \ref{fig1}c}. The 22 node features represent morphology attributes of the nucleus (7 features), the texture of the nucleus (7 features), a one-hot encoded vector of the nucleus class (6 features) and the nucleus centroid coordinates (2 features). The morphology features are: area, perimeter, eccentricity, solidity, major axis length, minor axis length, and extent. The texture features are: roughness, contrast, dissimilarity, homogeneity, entropy, angular second moment, and dispersion. Finally, we represent each class label as a one-hot encoded vector $\mathbf{c}_i \in \{0,1\}^{6}$, where exactly one element equals 1, indicating the class membership, and all remaining elements are equal to 0.

\subsection*{Implementation details of TILE-Graphs}

\begin{figure}[]
    {\centering
    \includegraphics[width=1\textwidth]{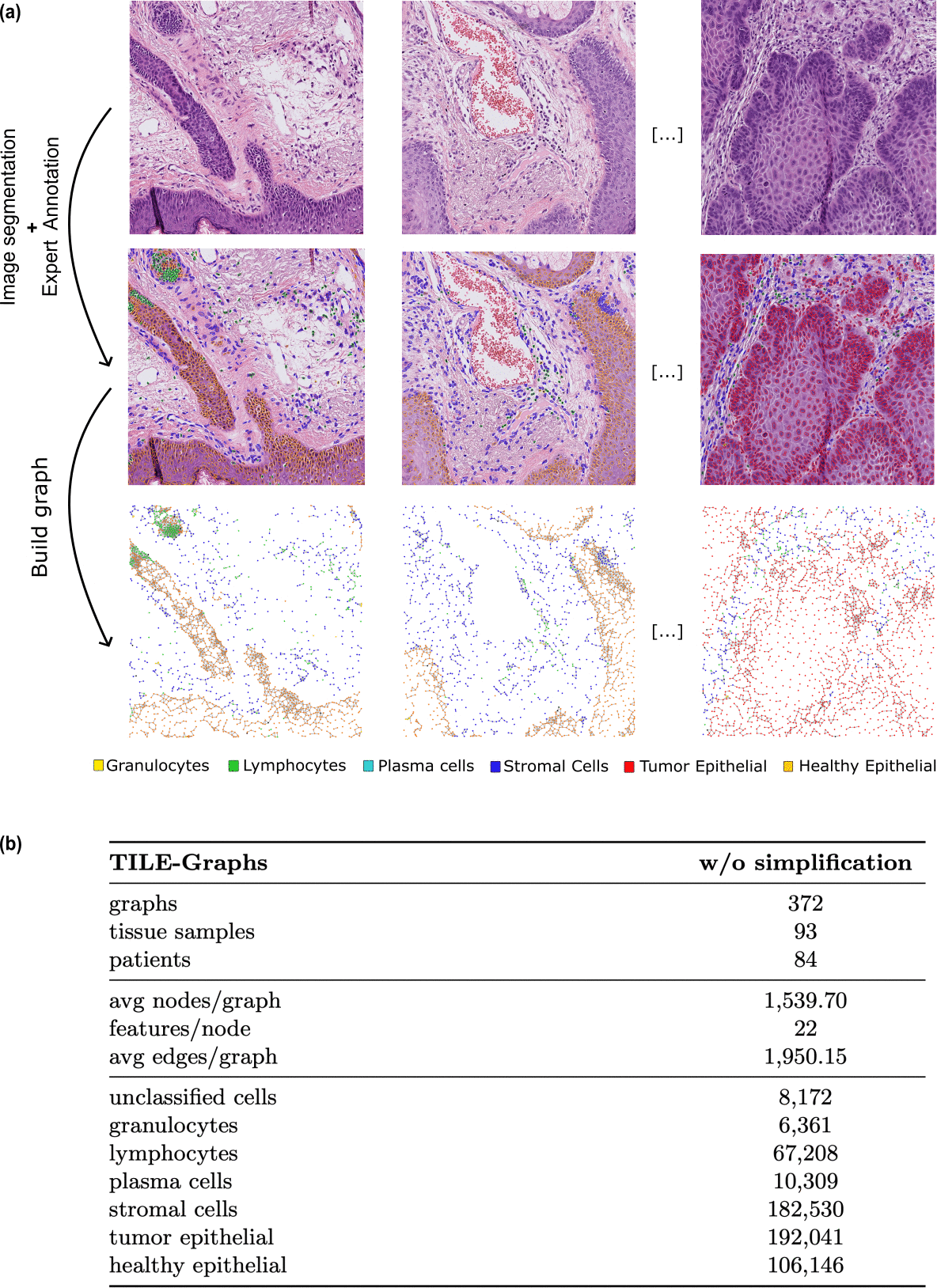}
    \par}
    \caption[Caption Title]{\small \textbf{Description of  TILE-Graphs.}
    \footnotesize \textbf{(a)} Steps to generate TILE-Graphs. First, patches from cSCC patient sample extracted from tumor epithelial and healthy epithelial regions are segmented using SCC Hovernet. Then from these segmentation maps 372 graphs are built. \textbf{(b)} TILE-Graphs dataset statistics. It includes 372 patches from 93 samples from 84 patients. The resulting 372 graphs are then split keeping graphs of the same patients in the same split during cross-validation.}
    \label{fig2}
\end{figure}

We collected 93 WSIs from 84 cSCC patients in which tumor regions were annotated by two experts (each expert having a subset of the data to annotate). These WSIs are a subset of already published TumSeg dataset \cite{Sancere2026}, where the 93 WSIs kept here contain both tumor and healthy epithelial tissue. Contrary to TumSeg where the images were downsampled, the collected WSIs are at 40x magnification. To generate TILE-Graphs we extracted two 2560x2560 pixel patches fully inside the annotated tumor regions and two 2560x2560 pixel patches containing healthy epithelial cells from each of the WSIs. We then built one graph per patch. \textbf{Figure \ref{fig2}a} shows the generation of 3 graphs from the dataset. Contrary to WSI-Graph dataset we did not simplify the graphs as they are smaller and we aim to explore their complexity for classification. Additionally, for these patches isolated epithelial cells are most probably not miss-classified, and then the simplification process would remove essential nodes. We used the same threshold distance to generate edges as in WSI-Graph dataset: $r_0=50\operatorname{pixels}$ corresponding to $r_0\approx \SI{11.5}{\micro\meter}$. A comprehensive description of TILE-Graphs is shown in \textbf{Figure \ref{fig2}b}. The node features are the same as in WSI-Graph.

\subsection*{Graph Transformer with linear complexity}
\label{graphTransformers}

In this work we perform binary classification of graph nodes as tumor or healthy epithelial cells. Standard Graph Transformers show high performance in node classification \cite{Kim2022,Ying2021,Rampasek2022}, but are computationally intensive due to the quadratic complexity of the attention mechanism relative to number of nodes. This restricts their practical application to small graphs only (e.g., on the order of hundreds of nodes) \cite{Rampasek2022}. \medskip   

In our work we evaluated NodeFormer \cite{Wu2022nodeformer}, DIFFormer \cite{Wu2023difformer} and SGFormer models \cite{Wu2023sgformer} - Graph Transformers with linear computational complexity w.r.t the number of nodes. In NodeFormer, the original softmax attention is approximated using stochastic kernel approximation, reducing the complexity to $\mathcal{O}(n)$. In DIFFormer, the exponential function in the softmax attention is replaced by its first-order Taylor expansion. This new attention layer can be efficiently computed using linear complexity via re-ordering the matrix product. SGFormer is composed of a one-layer global attention and a shallow GNN network. Contrary to all-pair attention that incurs $\mathcal{O}(n^2)$ complexity, this simple global attention allows for $\mathcal{O}(n)$ complexity. Here we developed an alternative training strategy to the originally implemented for these methods, which is described in the following sections.

\subsection*{Context-aware GNN classification of epithelial cells}
\label{training}

We evaluated state of the art GNNs and Graph Transformers with linear complexity for large graphs on $k=3$ max-hops-simplified graph $G^{(3)}$ for epithelial cell classification. Prior to training, continuous node features were standardized using z-score normalization computed from training set nodes and applied to both training and test sets. To prevent label leakage, the cell class features of epithelial nodes targeted for prediction were masked during training. Although the cell class feature was masked for epithelial nodes themselves, it remained available for neighboring nodes. Consequently, the graph model incorporates biologically meaningful contextual information, including neighboring cell class features, through message passing. The resulting epithelial node embeddings therefore depend on both intrinsic cellular features and features propagated from neighboring cells, enabling predictions based on the composition and spatial organization of the local tissue microenvironment. \medskip

In general form, node embeddings were computed as: 

\begin{equation}
\mathbf{H^{(3)}}_{(l+1)} =
\operatorname{\Phi_{(l)}} \left( \mathbf{H^{(3)}}_{(l)}, \mathbf{A^{(3)}} \right)
\end{equation}

where, $\mathbf{H^{(3)}}_{(l)} \in \mathbb{R}^{N \times d_l}$ denotes node embeddings at layer $l$ of $G^{(3)}$, $\mathbf{A^{(3)}} \in \mathbb{R}^{N \times N}$ is the adjacency matrix of $G^{(3)}$, and $\operatorname{\Phi_{(l)}(\cdot)}$  represents the architecture-specific propagation operator.  \medskip

All architectures were adapted to binary node classification by replacing the output layer with a single linear unit producing one logit and corresponding probability per node:

\begin{equation*}
\begin{aligned}
z_i &= w^\top h_i + b \\ \notag
\hat{y}_i &= \sigma(z_i) = \frac{1}{1 + e^{-z_i}} \notag
\end{aligned}
\end{equation*}

where $h_i \in \mathbb{R}^{d}$ denotes the learned embedding of node $i$, $w \in \mathbb{R}^{d}$ and $b \in \mathbb{R}$ are learnable parameters shared across nodes, $z_i \in \mathbb{R}$ is the predicted logit, $\hat{y}_i \in (0,1)$ is the predicted probability, and $\sigma(\cdot)$ denotes the sigmoid function. Model parameters were optimized using binary cross-entropy loss computed exclusively over epithelial nodes:

\begin{equation}
\mathcal{L} =
-
\sum_{i \in V_{\mathrm{epi}}}
\left[
y_i \log(\hat{y}_i)
+
(1 - y_i)\log(1 - \hat{y}_i)
\right]
\end{equation}

where $y_i \in \{0,1\}$ denotes the ground-truth label of node $i$ and $V_{\mathrm{epi}} \subseteq V$ denotes the set of epithelial nodes. Finally, all model parameters were optimized using Adam algorithm \cite{Kingma2014} to minimize  $\mathcal{L} $, with Adam parameters depending on model chosen.  \medskip  

To evaluate the model performance on binary node classification we ran 3-fold cross-validation using 2 folds for training and 1 for testing. We trained for 200 epochs on one 80GB A100 NVIDIA GPU (Ampere micro-architecture) per model, without early stopping nor hyperparameter search. To train SGFormer, NodeFormer, and DIFFormer on WSI-Graph, we used the hyperparameter configuration from their respective training on the large graph \textsf{ogbn-proteins} \cite{Hu2020}. To train on TILE-Graph, we used the hyperparameter configuration from the methods' respective training on the medium graph \textsf{Cora} \cite{Sen2008}. We applied the default hyperparameters in all other GNNs as in \cite{sgformer_github}.

\subsection*{Image-based approaches for epithelial cell classification}

To compare GNNs' performance with image-based models, we created WSI-Graph baseline dataset consisting on the WSI used to build the graph and its segmentation maps. This dataset contains over 36,000 segmented patches of 256x256 pixels. We trained CellViT256 model for 200 epochs on one 80GB A100 NVIDIA GPU to segment and classify epithelial cells as tumor or healthy. We trained Hovernet for 2 distinct steps \cite{Graham2019} of 50 epochs on two 80GB A100 NVIDIA GPUs following original implementation. Prior to training on baseline dataset, CellViT256 model was originally pretrained on 104 million 256×256px histological image patches from The Cancer Genome Atlas (TCGA) and Hovernet was originally pretrained on ImageNet \cite{Deng2009} and the resulting weights were taken as initialization. We did not apply early stopping and used the default hyperparameter sets. We either trained the models on all the WSI patches or on patches containing at least 90\% of epithelial cells. Keeping only detected cells, we calculated balanced accuracy of classification of epithelial cells in 3-fold cross validation. \medskip

We also generated TILE-Graphs baseline dataset. It consists of 37,200 annotated patches generated from the 372 patches (2560x2560 pixel) used to build the graphs of TILE-Graphs. We evaluated image-based models on TILE-Graphs baseline dataset following the same strategy as for WSI-Graph baseline dataset, but only keeping the previously found best performing model.

\section*{Experiments and Results}

Healthy epithelial cells (also called non-neoplastic epithelial) and tumor epithelial (neoplastic epithelial) show very similar morphologies and are challenging to discriminate from one another in cSCC WSIs. In practice, pathologists rely on the global tissue architecture, overall composition, and on the surrounding cells to differentiate healthy from tumor regions. In this context, image-based models which process individual patches separately and lack information on the broader tissue structure, are unable to reliably classify these two cell types \cite{Sancere2026}. Both convolutional neural network models, such as Hovernet \cite{Graham2019} or Transformers such as CellViT \cite{Horst2024} process WSIs cropped into a large number of small tiles. Each of the tiles represents only a small portion of the WSI and therefore does not contain the broader tissue context, leading to reduced performance. \medskip

\subsection*{Comparison of graph-based and image-based approaches for epithelial cell classification}
\label{benchmarking1}

We compared the performance of linear-complexity Graph Transformers on large graphs, i.e SGFormer \cite{Wu2023sgformer}, NodeFormer \cite{Wu2022nodeformer}, and DIFFormer \cite{Wu2023difformer} with conventional GNNs that can also scale to large graphs within a reasonable computational budget, including GCN \cite{Kipf2017}, GAT \cite{Velickovic2018}, SGC \cite{Wu2019}, SGC-MLP (identical to SGC but with the final linear layer replaced by an MLP) \cite{sgformer_github}, and SIGN \cite{Frasca2020}. To ensure consistency and reproducibility across models, we adapted all architectures to the binary node classification by replacing the final output layer with a single output unit producing one logit per node. We evaluated binary node classification performance with following two distinct strategies.  \medskip

First, consistent with common practice \cite{Wu2023sgformer,Wu2022nodeformer,Wu2023difformer} we randomly sampled all nodes from  $k=3$ max-hops simplified WSI-Graph (WSI-Graph$^{(3)}$) into 3 folds and applied 3-fold cross-validation. In this standard transductive evaluation protocol, randomly selected test nodes are often adjacent to training nodes, allowing their representations to be influenced by training data through neighborhood aggregation during message passing. This can lead to overly optimistic performance estimates when the goal is to assess generalization to independent or disjoint graph regions, in particular in biological tissues where cells of the same type are often grouped together. \medskip

To overcome this bias, we used 100 subgraphs originating from WSI-Graph$^{(3)}$ split and sample them into 3 folds. Each subgraph consists of non-overlapping parts of the WSI-Graph$^{(3)}$. We ran 3-fold cross-validation and predicted the class of all epithelial cells in the test set in each run. Node class inference in entire subgraphs instead of random nodes from one single large graph allows to test the method performance in a real-life setup, where a WSI needs to be annotated as a whole. The resulting balanced accuracy and standard error from both evaluations are shown in \textbf{Table \ref{tab:smallbenchmark1}}. Following previous findings on large graphs, SGFormer and DIFFormer are the best performing models. \medskip

\begin{table}[t]
\centering
\begin{subtable}{1\linewidth}
\centering
\setlength{\tabcolsep}{5pt}

\begin{tabularx}{0.8\linewidth}{@{}Xcc@{}}
\toprule
\textbf{Method} & \textbf{Subgraphs} & \textbf{Random Nodes} \\
\midrule
DIFFormer  & \textbf{85.2 $\pm$ 1.5} & 91.1 $\pm$ 0.1 \\
SGFormer   & \textbf{85.1 $\pm$ 2.5} & \textbf{94.9$ \pm$ 0.2} \\
SIGN       & 80.5 $\pm$ 2.8          & 84.8 $\pm$ 0.6 \\
GAT        & 80.4 $\pm$ 1.0          & 73.3 $\pm$ 2.0 \\
SGCMLP     & 80.4 $\pm$ 2.7          & 79.3 $\pm$ 0.4 \\
NodeFormer & 79.0 $\pm$ 0.5          & 85.0 $\pm$ 1.4 \\
SGC        & 76.8 $\pm$ 2.2          & 66.8 $\pm$ 3.1 \\
GCN        & 70.4 $\pm$ 2.0          & 84.6 $\pm$ 1.5 \\
\bottomrule
\end{tabularx}
\end{subtable}
\vspace{5pt} 
\caption[Caption Title]{\small \textbf{Comparison of GNN models on binary epithelial node classification performance on simplified WSI-Graph.}
\footnotesize Balanced accuracy (\%) is reported as mean ± standard error over 3-fold cross-validation under subgraph-based and random node evaluation protocols and  $k=3$ max-hops simplification for the graph ( WSI-Graph$^{(3)}$).}
\label{tab:smallbenchmark1}
\end{table}

 We observed that image-based models show lower accuracy compared to SGFormer and DIFFormer on the same dataset. Indeed, SGFormer and DIFFormer respectively yielded in 85.2 $\pm$ 1.5 and 85.1 $\pm$ 2.5 balanced accuracy ($\pm$ standard error) whereas best tested image-based approach CellViT256 yielded in 81.2 $\pm$ 3.0 balanced accuracy. We were unable to train CellViT-SAM-B model as, even with mixed-precision training, it could not fit into the memory of an 80GB A100 NVIDIA GPU with an acceptable batch size. Results are listed in \textbf{Table \ref{tab:smallbenchmark2}}. Interestingly, SGFormer was previously shown to outperform all GNNs of \textbf{Table \ref{tab:smallbenchmark2}} in node classification task \cite{Wu2023sgformer}, using a random nodes evaluation protocol. With subgraphs protocol evaluation, its performance is comparable to DIFFormer.

\begin{table}[t]
\centering
\begin{subtable}{1\linewidth}
\centering
\setlength{\tabcolsep}{5pt}

\begin{tabularx}{0.8\linewidth}{@{}Xcc@{}}
\toprule
\textbf{Method} & \textbf{Training set} & \textbf{3-fold crossval} \\
\midrule
Hovernet      & all patches     & 73.7 $\pm$ 1.4 \\
    & epithelial only & 79.3 $\pm$ 2.3 \\
CellViT256    & all patches     & 76.5 $\pm$ 1.9 \\
    & epithelial only & \textbf{81.2 $\pm$ 3.0} \\
CellViT-SAM-B & all patches     & OOM \\
   & epithelial only & OOM \\
\bottomrule
\end{tabularx}
\end{subtable}
\vspace{5pt} 
\caption[Caption Title]{ \small \textbf{Comparison of image-based models on binary epithelial cell classification performance on WSI images.}
\footnotesize Balanced accuracy (\%) is reported as mean ± standard error over 3-fold cross-validation for different methods and training set configurations, using image-based representations.}
\label{tab:smallbenchmark2}
\end{table}

\subsection*{WSI-Graph node features}

We conducted a feature ablation study on  WSI-Graph$^{(3)}$ to evaluate the impact of node features on binary node classification. We evaluated SGFormer using the protocols described previously. We tested the impact of different node feature combinations on performance. Nucleus coordinates were part of all the feature combinations that we tested. We computed z-score normalization for each feature on the training set and applied to the test set using the same normalization parameters. We also tested the impact of such normalization on the performance. The results are reported in \textbf{Table \ref{tab:nodefeat}} and show that removing any type of feature and normalization leads to decrease in performance, highlighting their crucial contribution to model generalization.

\begin{table}[t]
\centering
\begin{subtable}{1\linewidth}
\centering
\setlength{\tabcolsep}{5pt}
\begin{tabularx}{\linewidth}{@{}Xccc@{}}
\toprule
\textbf{Node Features} & \textbf{z-score norm} & \textbf{Subgraphs} & \textbf{Random Nodes} \\
\midrule
morphology                           & no    & 67.8 $\pm$ 3.6          & 92.6 $\pm$ 0.3 \\
morphology                           & yes   & 79.6 $\pm$ 2.4          & 94.0 $\pm$ 0.5 \\
morphology \& texture                & no    & 70.6 $\pm$ 9.0          & 94.0 $\pm$ 0.9 \\
morphology \& texture                & yes   & 84.2 $\pm$ 1.6          & 94.3 $\pm$ 0.6 \\
morphology \& cell class             & no    & 73.6 $\pm$ 3.7          & 93.7 $\pm$ 0.5 \\
morphology \& cell class             & yes   & 84.0 $\pm$ 2.8          & 94.5 $\pm$ 0.4 \\
morphology \& texture \& cell class  & no    & 71.4 $\pm$ 8.4          & 92.8 $\pm$ 1.1 \\
morphology \& texture \& cell class  & yes   & \textbf{85.1 $\pm$ 2.5} & \textbf{94.9 $\pm$ 0.2} \\
\bottomrule
\end{tabularx}
\end{subtable}
\vspace{5pt} 
\caption[Caption Title]{ \small \textbf{Impact of node features on binary node classification performance on simplified WSI-Graph.}
\footnotesize Balanced accuracy (\%) is reported as mean ± standard error over 3-fold cross-validation for different node features under subgraph and random node evaluation protocols with SGFormer model and WSI-Graph$^{(3)}$.}
\label{tab:nodefeat}
\end{table}

\subsection*{WSI-Graph simplification}

To evaluate the impact of graph simplification of WSI-Graph on binary node classification performance, we progressively restricted the graph connectivity by limiting the maximum number of hops between nodes. This simplification reduces long-range connections and constrains information propagation during message passing. We evaluated SGFormer performance using the evaluation protocols described previously. \medskip

As shown in \textbf{Table \ref{tab:graphvar}}, graph simplification has a noticeable impact on performance. The best balanced accuracy is achieved with a 10-hop simplification ($86.6 \pm 2.2$), while both more restrictive configurations (1–2 max-hops) and the absence of simplification result in lower performance. This suggests that an intermediate level of connectivity provides sufficient contextual information for accurate classification while avoiding the propagation of less relevant signals. In contrast, performance under random node splits remains largely stable across all simplification levels. These results highlight that local graph structure contains most of the useful information for node classification, and that moderate graph simplification can improve robustness while reducing graph complexity. \medskip

A max-hops simplification with $k=3$ provides a good compromise between graph sparsity and model performance; therefore, this simplification was adopted throughout this work.

\begin{table}[t]
\centering
\begin{subtable}{1\linewidth}
\centering
\setlength{\tabcolsep}{5pt}
\begin{tabularx}{\linewidth}{@{}Xcc@{}}
\toprule
\textbf{Graph Simplifications}  & \textbf{Subgraphs} & \textbf{Random Nodes} \\
\midrule
No simplification  & 82.2 $\pm$ 2.9           & 94.6 $\pm$ 0.1 \\
50 max-hops        & 83.3 $\pm$ 3.3           & 94.9 $\pm$ 0.4 \\
10 max-hops        & \textbf{86.6 $\pm$ 2.2}  & 95.0 $\pm$ 0.2 \\
5 max-hops         & 82.5 $\pm$ 1.9           & 94.6 $\pm$ 0.5 \\
4 max-hops         & 82.1 $\pm$ 4.3           & 95.0 $\pm$ 0.4 \\
3 max-hops         & 85.1 $\pm$ 2.5           & 94.9 $\pm$ 0.2 \\
2 max-hops         & 81.8 $\pm$ 0.8           & \textbf{95.4 $\pm$ 0.3} \\
1 max-hops         & 81.1 $\pm$ 3.9           & 94.7 $\pm$ 0.6 \\
\bottomrule
\end{tabularx}
\end{subtable}
\vspace{8pt} 
\caption[Caption Title]{ \small \textbf{Impact of graph simplification on binary node classification performance on simplified WSI-Graph.}
\footnotesize Balanced accuracy (\%) is reported as mean ± standard error over 3-fold cross-validation for different maximum hop thresholds under subgraph and random node evaluation protocols with SGFormer model.}
\label{tab:graphvar}
\end{table}

\subsection*{Comparison of graph-based and image-based approaches on multiple patients dataset}

In the WSI-Graph approach described above we compared the performance of GNN with image-based model using a large graph derived from a WSI of a single patient. Evaluating CellViT on complete WSIs from different patients is computationally prohibitive and thus prevented a systematic training and evaluation on an entire WSIs dataset. We therefore devised another approach to enable the evaluation of our approach across a larger number of patients. Following this goal we used TILE-Graphs and its baseline counterpart, including samples from 84 patients, to evaluate both CellViT256 and GNNs in 3-fold cross validation as previously described. Samples from the same patient were assigned to the same fold, and all models were tested using the same splits. \medskip

Our results show that DIFFormer outperforms vision Transformer CellViT256 on the binary classification of epithelial cells as healthy or tumor (\textbf{Table \ref{tab:tilegraph}}). DIFFormer model resulted in classification with 83.6 $\pm$ 1.9 balanced accuracy on 3 fold cross-validation and CellViT256 with 78.1 $\pm$ 0.5 balanced accuracy with the same folds. Interestingly, SGFormer performed poorly on smaller graphs, even with adjusted hyperparameters. Indeed, its very light Graph Transformer architecture probably resulted in attention focused on very few nodes, and in the case of smaller graphs, these nodes might not represent enough the overall graph structure.

\begin{table}[t]
\centering
\begin{subtable}{0.8\linewidth}
\centering
\setlength{\tabcolsep}{5pt}

\begin{tabularx}{0.8\linewidth}{@{}Xcc@{}}
\toprule
\textbf{Method} & \textbf{3-fold crossval}  \\
\midrule
CellViT256 & 78.1 $\pm$ 0.5  \\
\midrule
\midrule
DIFFormer  & \textbf{83.6 $\pm$ 1.9}  \\
NodeFormer & 78.7 $\pm$ 4.4 \\
SGCMLP     & 66.4 $\pm$ 1.1 \\
SIGN       & 66.3 $\pm$ 1.0  \\
SGC        & 63.8 $\pm$ 1.2 \\
GCN        & 61.9 $\pm$ 0.9  \\
GAT        & 61.6 $\pm$ 1.4  \\
SGFormer   & 61.0 $\pm$ 0.8 \\
\bottomrule
\end{tabularx}
\end{subtable}
\vspace{8pt} 
\caption[Caption Title]{\small \textbf{Comparison of graph-based and image-based models on binary epithelial cell classification performance on TILE-Graphs and baseline dataset.}
\footnotesize Balanced accuracy (\%) is reported as mean ± standard error over 3-fold cross-validation for different methods and training set configurations.}
\label{tab:tilegraph}
\end{table}

\section*{Discussion}

WSIs of cancer samples contain rich information for medical diagnosis and for gaining insights into tumor biology. These images contain thousands to million of cells of different types whose automated segmentation and classification can help practitioners in comprehensive assessment of tumor samples. Due to the large size of WSIs, segmentation models are trained on small image patches. Within small patches the overall tissue structure and organization are missing. Here we focus on the problem of discrimination between healthy and tumor epithelial cells in cSCC. These cells show very similar morphologies and are impossible to discern within small patches. For this reason, pathologists rely on the global tissue architecture, overall composition, and on the surrounding cells to differentiate healthy tissue from tumor regions. To address this limitation of the WSI segmentation models, we explored, the use of GNNs, for incorporation of broader tissue context to differentiate healthy epithelial from tumor epithelial cells. We showed that scalable Graph Transformer architectures outperform image-based approaches in this task.  \medskip

In this work we show that creating graphs from histopathology images of cSCC, where each node represent a single cell, and using scalable Graph Transformers for epithelial cell classification outperforms the image-based approaches in this task. We first evaluated several graph-based and image-based models on the same annotated WSI and found that Graph Transformer models SGFormer and DIFFormer outperform image-based approach CellViT256. To further evaluate both graph-based and image-based models on multiple patients, we collected TILE-Graphs dataset. This dataset is composed of 372 patches from 93 WSIs of 84 different patients. The patches are converted into graphs and used to train GNNs. Again, a graph-based approach, DIFFormer, showed a superior classification performance compared to CellViT256. The higher accuracy of graph models in classifying the two cell types suggests that broader tissue context, encoded in graphs is indeed important for this task and that graph offers a convenient way to represent it. \medskip

In addition to performance, an important advantage of graph based models is their lower computational cost. Graph representation of tissues is computationally lighter compared to raw images and can be manipulated with ease. On the TILE-Graphs image baseline dataset, training CellViT256 required approximately five days to complete a single run of 3-fold cross-validation on a 80GB A100 NVIDIA GPU but only 31 min 59 s for graph architecture such as DIFFormer. Graph-based approaches for cell classification represent therefore a powerful and efficient way for WSI analysis compared to traditional computer vision methods.  \medskip  

Graph construction for the TILE-Graphs and WSI-Graph datasets requires initial segmentation inference using a pre-trained cSCC Hovernet model. No retraining of the model is needed thereby minimizing computational overhead. In addition, the feature ablation study exhibited that comparable node classification performance can be reached without the class feature of the neighboring nodes. Panoptic segmentation from cSCC Hovernet inference could then be skipped to further reduce computational cost, keeping only instance segmentation step.   \medskip  

In the present study, tissue was represented as a simple undirected graph in which nodes were associated with handcrafted morphological and spatial features. Future approaches could leverage pretrained foundation models for cancer WSIs, such as VOLTA \cite{Nakhli2024} to derive informative learned representations of cellular phenotypes and incorporate them as node attributes. Furthermore, more expressive graph formalisms, including multi-graphs or hypergraphs, could be explored to capture higher-order and multi-relational interactions between cells through alternative edge or hyperedge construction strategies.

\clearpage

\section*{Data availability}

WSI-Graph and TILE-Graphs datasets as well as its image baseline datasets counterparts are published within a Zenodo repository: \href{https://doi.org/10.5281/zenodo.21415205}{https://doi.org/10.5281/zenodo.21415205}. 

\section*{Code availability}

The github repository linked to this project is available at \href{https://github.com/bozeklab/wsi-graph-formers}{https://github.com/bozeklab/wsi-graph-formers}. 

\section*{Competing Interest}

Authors declare no conflicts of interests.

\section*{Author Contributions}

\noindent\textbf{Conceptualization}: Lucas Sancéré, Noémie Moreau, Katarzyna Bozek. \newline
\textbf{Data Curation}: Lucas Sancéré. \newline
\textbf{Formal Analysis}: Lucas Sancéré.   \newline
\textbf{Funding Acquisition}: Katarzyna Bozek.   \newline
\textbf{Investigation}: Lucas Sancéré. \newline
\textbf{Methodology}: Lucas Sancéré. \newline
\textbf{Project Administration}: Lucas Sancéré, Noémie Moreau, Katarzyna Bozek. \newline
\textbf{Resources}: Lucas Sancéré.   \newline
\textbf{Software}:  Lucas Sancéré. \newline
\textbf{Supervision}: Noémie Moreau, Katarzyna Bozek.  \newline
\textbf{Validation}:  Lucas Sancéré. \newline
\textbf{Visualization}:  Lucas Sancéré.  \newline 
\textbf{Writing – Original Draft Preparation}:  Lucas Sancéré, Noémie Moreau, \newline Katarzyna Bozek. \newline
\textbf{Writing – Review \& Editing}: Lucas Sancéré, Noémie Moreau, \newline Katarzyna Bozek.

\bibliographystyle{naturemag} 
\bibliography{refs} 

\end{document}